%% file: main.tex
\documentclass[runningheads]{llncs}

\usepackage{eccv}

\usepackage{eccvabbrv}

\usepackage{graphicx}
\usepackage{wrapfig}
\usepackage{booktabs}
\usepackage{colortbl}

\usepackage[accsupp]{axessibility}
\usepackage{multirow}
\usepackage{makecell}
\usepackage{marvosym}

\usepackage[breaklinks,colorlinks,citecolor=eccvblue]{hyperref}

\usepackage{orcidlink}

\newcommand{\sysname}{\textit{Diff-ES}\xspace}
\newcommand{\ourscell}[1]{\cellcolor{eccvblue!10}#1}
\newcommand{\corrmark}{\textsuperscript{\Letter}}
\begin{document}
\crefname{appendix}{Appendix}{Appendices}
\Crefname{appendix}{Appendix}{Appendices}

\title{Diff-ES: Stage-wise Structural Diffusion Pruning via Evolutionary Search}

\author{Zongfang Liu\inst{1,2}\thanks{Equal contribution.} \and
Shengkun Tang\inst{3}\textsuperscript{$\star$} \and
Zongliang Wu\inst{1,2} \and
Xin Yuan\inst{2}\corrmark \and
Zhiqiang Shen\inst{3}\corrmark}

\authorrunning{Z. Liu et al.}

\institute{Zhejiang University \and
Westlake University \and
Mohamed bin Zayed University of Artificial Intelligence}

\maketitle

\input{sec/0_abstract}

\input{sec/1_intro}
\input{sec/2_relatedwork}

\input{sec/3_method}
\input{sec/4_experiments}

\input{sec/5_conclusion}
\bibliographystyle{splncs04}
\bibliography{main}
\clearpage
\appendix
\crefalias{section}{appendix}
\renewcommand{\thesection}{\Alph{section}}
\renewcommand{\theHsection}{appendix.\Alph{section}}
\input{sec/6_suppl}

\end{document}

%% file: sec/0_abstract.tex
\begin{abstract}
Diffusion models have achieved remarkable success in high-fidelity image generation but remain computationally demanding due to their multi-step denoising process and large model sizes. Although prior work improves efficiency either by reducing sampling steps or by compressing model parameters, existing structured pruning approaches still struggle to balance real acceleration and image quality preservation. 
In particular, prior methods such as MosaicDiff rely on heuristic, manually tuned stage-wise sparsity schedules and stitch multiple independently pruned models during inference, which increases memory overhead. However, the importance of diffusion steps is highly non-uniform and model-dependent. As a result, schedules derived from simple heuristics or empirical observations often fail to generalize and may lead to suboptimal performance.
To this end, we introduce \textbf{\sysname}, a stage-wise structural \textbf{Diff}usion pruning framework via \textbf{E}volutionary \textbf{S}earch, which optimizes the stage-wise sparsity schedule and executes it through memory-efficient weight routing without model duplication. \sysname divides the diffusion trajectory into multiple stages, automatically discovers an optimal stage-wise sparsity schedule via evolutionary search, and activates stage-conditioned weights dynamically without duplicating model parameters.  Our framework naturally integrates with existing structured pruning methods for diffusion models including depth and width pruning.
Extensive experiments on DiT and SDXL demonstrate that \sysname{} consistently achieves wall-clock speedups while incurring minimal degradation in generation quality, establishing state-of-the-art performance for structured diffusion model pruning. Codes are available at \url{https://github.com/ZongfangLiu/Diff-ES}.

\keywords{Diffusion Model Compression \and Structural Pruning \and Evolutionary Search \and Stage-wise Sparsity Schedule \and Weight Routing}
\end{abstract}

%% file: sec/1_intro.tex
\section{Introduction}
\label{sec:intro}
\begin{figure}
    \centering
    \includegraphics[width=0.5\linewidth]{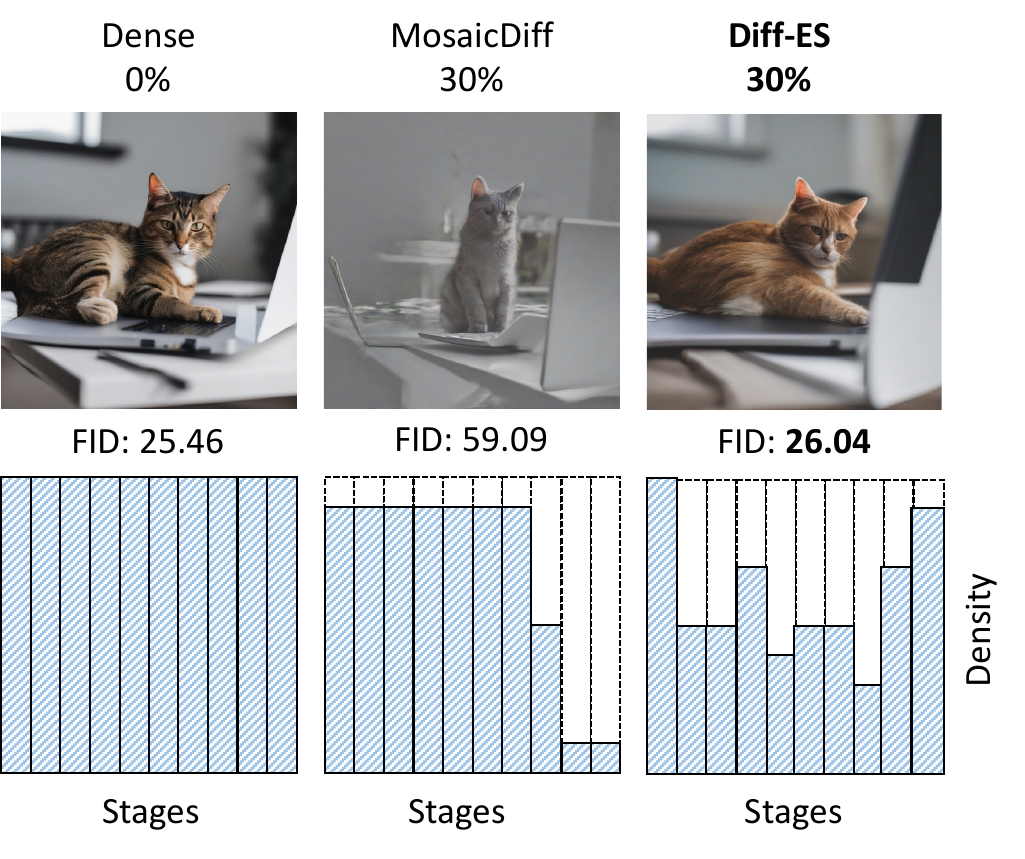}
    \caption{\textbf{Comparison of sparsity ($1 - \text{density}$) schedules.} An effective stage-wise sparsity schedule is crucial for maintaining image quality. Using the same structural second-order pruning method, our \sysname framework significantly outperforms MosaicDiff on SDXL by employing an optimized, adaptive stage-wise sparsity schedule. For fair comparison, the original 3-stage schedule of MosaicDiff is visualized in 10.}
    
    \label{fig:motivation}
\end{figure}

Diffusion models~\cite{peebles2023scalable, podell2023sdxl} have become the mainstream techniques for high-fidelity image generation, including various tasks such as unconditional generation~\cite{song2020denoising}, text-guided generation~\cite{rombach2022high, podell2023sdxl} and image editing~\cite{meng2021sdedit, kawar2023imagic}. However, deploying diffusion models remains challenging due to their high computational requirements, especially in real-time generation scenarios.

The generation speed of diffusion models is impacted by two primary factors: the multi-step nature of iterative denoising and the rapidly increasing backbone model size. 
To improve efficiency, prior work has pursued two largely orthogonal directions:  reducing the sampling steps \cite{song2020denoising, lu2022dpm, song2023consistency} or the computation cost in each sampling step \cite{fang2023structural, ma2024deepcache, guo2025mosaicdiff, zhu2025obs}. In this work, we focus on the latter strategies including structured width and depth pruning (layer dropping). While the previous work obtains the same compressed model for all sampling steps via structured pruning, recent works reveal that the computation resource allocated for different sampling steps can be varied 
\cite{li2023autodiffusion, pan2024t, guo2025mosaicdiff, tang2024adadiff}.
Among them, MosaicDiff~\cite{guo2025mosaicdiff}, which partitions sampling into three coarse stages, sets stage-wise structured pruning manually, and stitches three independently pruned models during sampling. While compatible with various structured techniques, MosaicDiff’s stage division and computation allocation rely heavily on heuristics and manual tuning, which may result in suboptimal results, as illustrated in \Cref{fig:motivation}. This motivates us to ask: \textit{Can we develop an approach to automatically allocate computational budget across sampling steps?}

\begin{figure}[t]
  \centering
  \includegraphics[width=0.98\textwidth,page=1]{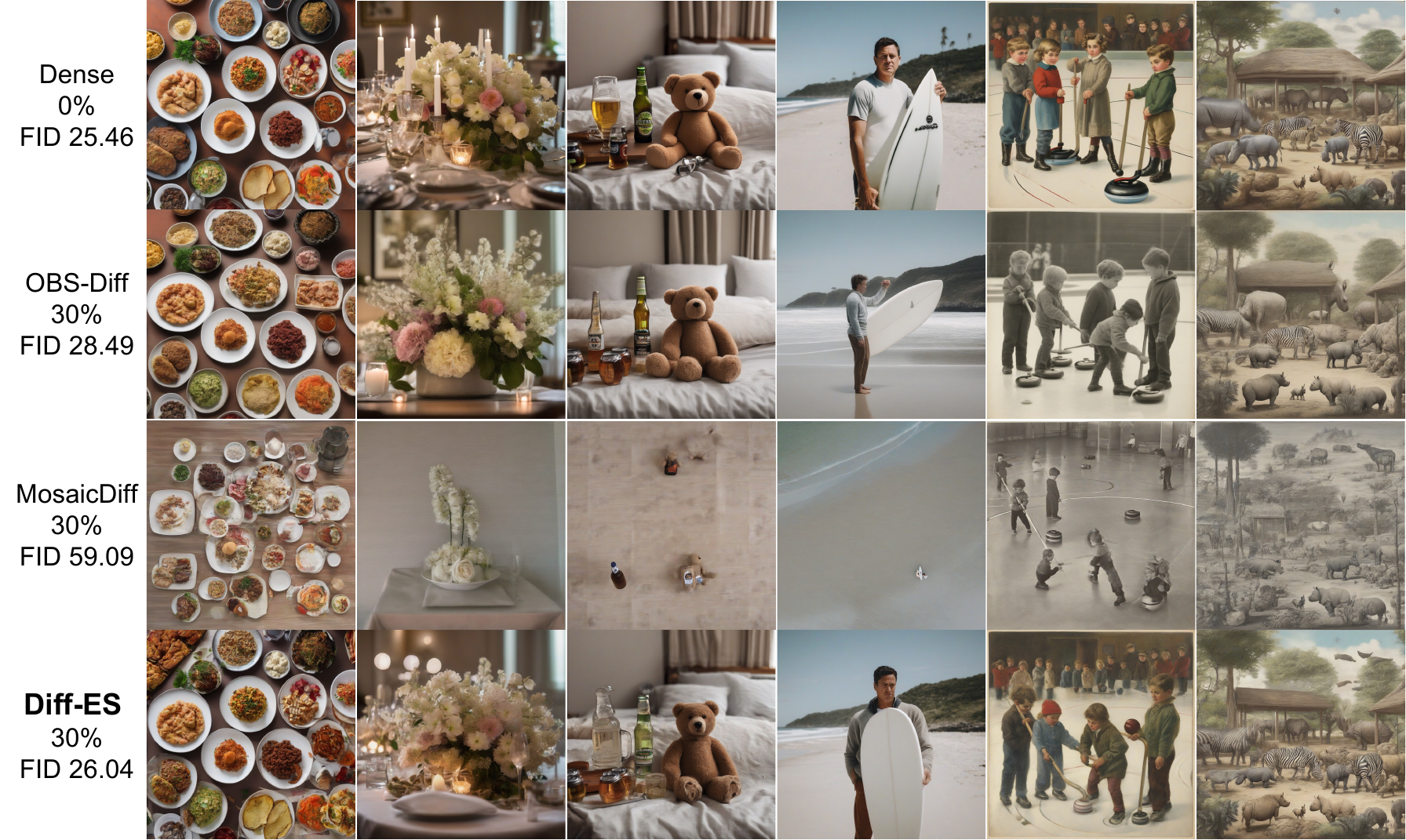}
\caption{\textbf{Visual comparison on SDXL-base-1.0 under 30\% sparsity.}
Each row shows generations from the same prompts using 20 sampling steps (CFG 7.5): Dense, OBS-Diff, MosaicDiff with OBS, and \sysname\ with OBS.
At the same sparsity level, \sysname\ remains closest to the dense model, preserving object identity, scene layout, and fine textures. OBS-Diff already shows clear structural differences with dense model (e.g., the teddy bear is rendered with three legs), while MosaicDiff exhibits much more severe semantic and perceptual degradation. This visual trend is consistent with FID.}
  \label{fig:examples}
\end{figure}

To bridge this gap, we propose \sysname, which optimizes stage-wise sparsity under a fixed global budget via evolutionary search.
Given a target sparsity, we partition the denoising trajectory into several stages (i.e. 10, 20) and initialize a population using uniform, random, and patterned schedules.
At each generation, offspring are produced by mutating sparsity levels of selected stages while preserving the global budget, and candidates are ranked by a lightweight fitness metric.
Top candidates survive to the next generation, and the mutation–evaluation–selection cycle is repeated for a fixed number of generations, yielding the final stage-wise sparsity schedule.
As a plug-and-play framework, \sysname is compatible with both width pruning (Wanda/OBS) and depth pruning (LayerDrop), and uses lightweight weight routing to avoid model stitching as required in MosaicDiff.
We validate our method combined with Layer Dropping, Wanda, and second-order structural pruning (OBS) on both CNN-based model (SDXL) and Transformer-based model (DiT). \sysname achieves superior image quality compared to all baselines, including Diff-Pruning~\cite{fang2023structural}, DeepCache~\cite{ma2024deepcache}, and OBS-Diff~\cite{zhu2025obs}. Across different pruning techniques, \sysname consistently surpasses MosaicDiff in all settings. \Cref{fig:examples} illustrates that \sysname preserves strong semantic alignment and perceptual quality.

\noindent We summarize our main contributions as below:
\begin{itemize}
    \item We identify a key limitation of heuristic stage-wise structural pruning for diffusion models (e.g., MosaicDiff), where manually tuned coarse schedules fail to capture model-dependent sparsity patterns and consequently generalize poorly across architectures; we therefore propose a novel evolutionary framework for optimizing stage-wise sparsity schedules.

    \item \sysname serves as a general framework with plug-and-play compatibility for structured pruning methods such as Layer Dropping, Wanda, and OBS.

    \item  We conduct comprehensive experiments to demonstrate the effectiveness of \sysname. We evaluate \sysname on both CNN-based (SDXL) and Transformer-based (DiT) diffusion models. Across all settings, \sysname consistently outperforms strong baselines such as Diff-Pruning, OBS-Diff, and MosaicDiff.
\end{itemize}

%% file: sec/2_relatedwork.tex
\section{Related Work}
\label{sec:relatedwork}
\paragraph{\textbf{Efficient Diffusion Models.}}
Improving the efficiency of diffusion models has been actively investigated from both algorithmic and architectural perspectives. On the algorithmic side, acceleration is typically achieved by reducing the number of denoising steps through advanced ODE/SDE solvers or distillation-based compression. Non-Markovian solvers such as DDIM~\cite{song2020denoising}, DPM-Solver~\cite{lu2022dpm}, and Consistency Models~\cite{song2023consistency} enable few-step or even single-step generation, while distillation frameworks~\cite{meng2023distillation,salimans2022progressive} compress long-horizon denoising behavior into compact student models.
Architectural approaches, in contrast, focus on lowering the computation cost within each denoising step. Structural pruning removes redundant channels and layers in diffusion backbones~\cite{fang2023structural,kim2023architectural}, and TinyFusion~\cite{fang2025tinyfusion} introduces depth pruning to streamline DiT architectures. EcoDiff~\cite{zhang2024effortless} presents a general pruning framework applicable to diverse text-to-image diffusion models. Complementary methods further improve backbone efficiency through architectural compression and model distillation~\cite{kim2024bk,meng2023distillation}, while dynamic and adaptive strategies exploit temporal heterogeneity in diffusion trajectories via early-stopping~\cite{lyu2022accelerating} or adaptive computation scheduling~\cite{tang2024adadiff}.
Additional techniques leverage quantization for lightweight inference~\cite{he2023ptqd,li2024svdquant,li2023q,shang2023post} and feature caching to reuse intermediate activations across timesteps~\cite{ma2024learning,ma2024deepcache,zhu2024dip}, and besides, trajectory stitching~\cite{pan2024t} integrates models of varying complexities across inference stages to balance efficiency and fidelity. AutoDiffusion \cite{li2023autodiffusion} employs evolutionary search to optimize timestep usage and layer skipping, yet its reliance on costly FID-based fitness makes it impractical for large-scale diffusion models, and its naive layer-drop strategy limits achievable sparsity and degrades image quality.

\paragraph{\textbf{Post-training Pruning Methods.}}
Early post-training compression methods, including OBD and OBS, adopted Hessian-based saliency for weight pruning \cite{lecun1989optimal,hassibi1993optimal}. However, estimating the full Hessian is often prohibitively expensive, motivating layer-wise approximations such as OBC \cite{frantar2022optimal}. In large language models, SparseGPT formulates pruning as a layer-local sparse regression using approximate second-order information, achieving $\sim$50\% unstructured sparsity in one shot \cite{frantar2023sparsegpt}. Wanda \cite{sun2023simple} further simplifies pruning via activation-weight magnitude heuristics, matching similar sparsity without post weight-updating for compensation. Subsequently, OBS has been extended to structural granularity \cite{kurtic2023ziplm, ling2024slimgpt, wei2024structured, tang2025darwinlm}. Diffusion models are particularly sensitive to pruning due to the iterative denoising process that amplifies small perturbations \cite{ramesh2024efficient}. Early structural approaches \cite{fang2023structural, zhu2024dip} require retraining to maintain fidelity. Recent works \cite{zhang2024effortless, zhu2025obs} enable training-free structural pruning for diffusion models. However, they prune the same model across all diffusion timesteps, which overlooks the inherent step-wise importance dynamics in the diffusion sampling process, potentially missing further efficiency gains. MosaicDiff \cite{guo2025mosaicdiff} considers the step-wise dynamics by manually dividing the diffusion process into three stages and uses structural OBS to prune different models for different stages. Though insightful, the empirical hand-crafted stage division and sparsity allocation hinder their performance and generalizability.

%% file: sec/3_method.tex
\section{Method}

\begin{figure*}[t]
    \centering
    \includegraphics[width=\linewidth]{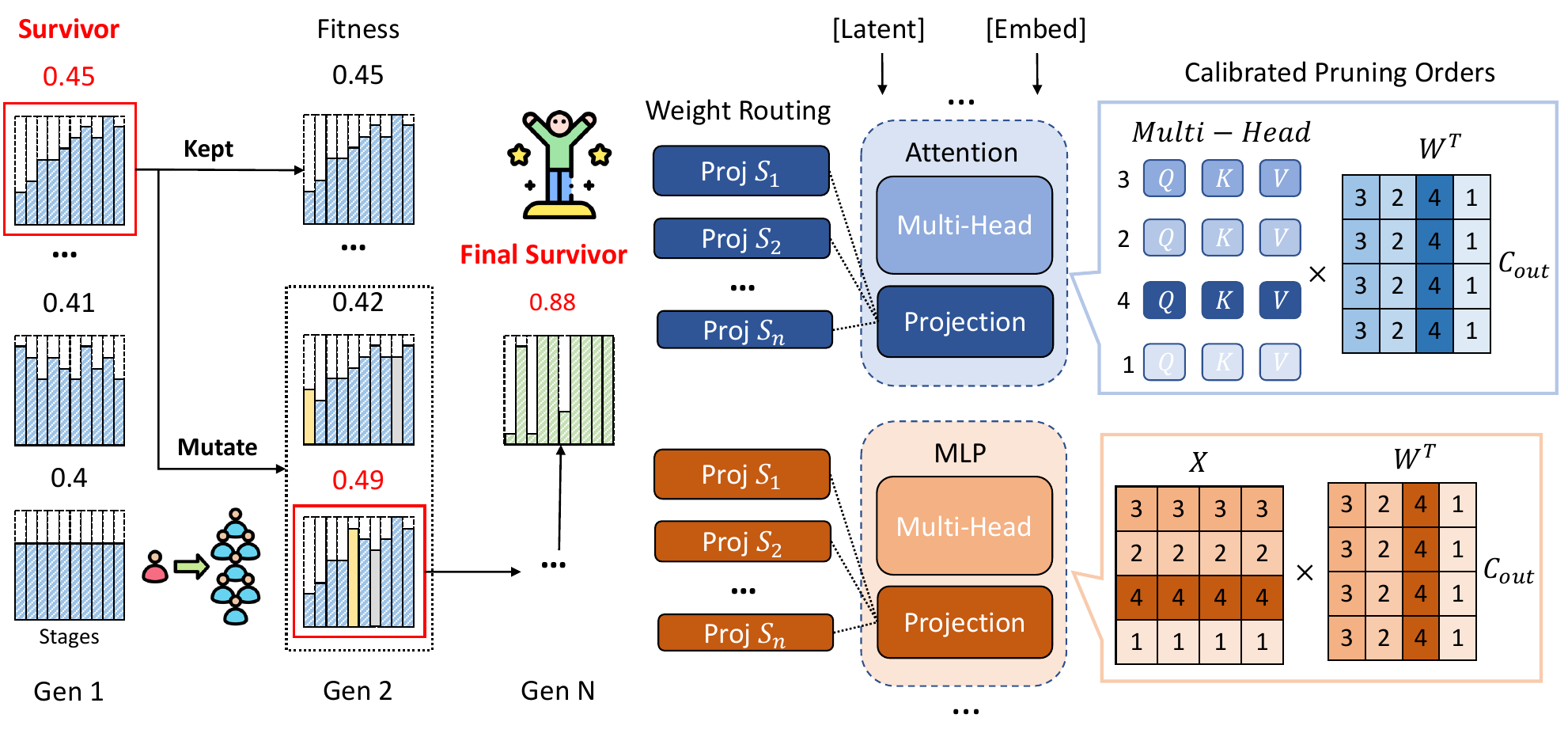}
    \caption{\textbf{Overview of \sysname.} We divide the diffusion process into stages and evolve an optimal stage-wise sparsity schedule under a fixed global budget. A level-switch mutation redistributes sparsity across stages, while lightweight fitness evaluation (e.g., TOPIQ/CLIP-IQA/SSIM) guides survivor selection. For each stage, SNR-aware calibration supports pruning methods to obtain calibrated projection orders. For methods with weight updating such as OBS, all updated weights are stored once in a compact database and retrieved during evaluation through our efficient weight-routing mechanism, enabling rapid model assembly without recomputing second-order updates.}
    \label{fig:framework}
\end{figure*}

\paragraph{\textbf{Framework Overview.}}
We aim to learn an \emph{optimal stage-wise sparsity schedule} for diffusion models under a fixed global budget by coupling evolutionary search with structural pruning methods such as OBS. 
As shown in \Cref{fig:framework}, we first divide the denoising process into $n$ stages and assign each stage an initial sparsity level to form a population of candidate sparsity schedules (\S\ref{sec:problem_setup}). An evolutionary algorithm with \emph{level-switch} mutation then explores these schedules while maintaining the global sparsity constraint (\S\ref{sec:evolutionary_search}). 
To support this process, we construct an \emph{SNR-aware} calibration set and use it to determine each stage’s pruning order via OBS (\S\ref{sec:stagewise_secondorder}), storing the resulting updated weights in a compact database. Finally, we introduce a lightweight \emph{weight routing mechanism} that retrieves these precomputed pruning trajectories, enabling rapid and memory efficient model assembly and fitness evaluation during search without recomputing second-order updates.

\subsection{Problem Setup}
\label{sec:problem_setup}
The diffusion process exhibits an inherently uneven importance across its denoising stages: early steps primarily construct global semantic structures, while later steps refine fine-grained visual details \cite{liew2022magicmix}. 
Motivated by this phenomenon, we argue that applying an identical pruning configuration to all stages is inefficient, as it neglects the stage-dependent contribution to generation quality. 
Therefore, instead of pruning a single shared model for the entire process, similar to \cite{guo2025mosaicdiff}, we focus on learning how to assign distinct sparsity levels to different diffusion stages. This stage-wise sparsity schedule better aligns computational resources with the functional significance of each stage, leading to more effective pruning under the same overall sparsity constraint.
Formally, let the diffusion process be divided into $n$ stages, denoted as $\{S_1, S_2, \ldots, S_n\}$, and let each stage $S_i$ be associated with a sparsity level $L_i$ that determines its pruning strength. 
Let $L_t$ denote the target overall sparsity ratio, $f(\mathbf{L})$ be a function evaluating model performance under a stage-wise sparsity schedule $\mathbf{L} = \{L_1, L_2, \ldots, L_n\}$, and $G(\mathbf{L})$ denote the resulting global sparsity. 
The optimization objective can then be expressed as:
\begin{equation}
   \hat{\mathbf{L}} = \textstyle \arg\max_{\mathbf{L}} f(\mathbf{L}) \quad \text{s.t.} \quad G(\mathbf{L}) \geq L_t. 
\end{equation}
This defines a high-dimensional, discrete, and non-differentiable optimization problem, as the mapping from $\mathbf{L}$ to model performance involves pruning and re-evaluation of the diffusion network. 
Therefore, gradient-based optimization is infeasible. To address this challenge, we employ an \emph{evolutionary search} strategy that evolves a population of sparsity schedules toward higher fitness. This approach naturally accommodates discrete operations.

\subsection{Level-Switch Evolutionary Search}
\label{sec:evolutionary_search}

\paragraph{\textbf{Initialization.}}
We initialize a population of candidate sparsity schedules, denoted by $\{L_1, L_2, \ldots, L_n\}$, each satisfying the target sparsity constraint $G(\mathbf{L}) = L_t$. 
Each individual represents a complete stage-wise sparsity schedule across the $n$ diffusion stages. 
The initial population can be generated using uniform or random sampling, or initialized with heuristic sparsity schedules.

\paragraph{\textbf{Fitness Evaluation.}}
Each candidate configuration $\mathbf{L}$ is evaluated by generating a small batch of images under the corresponding pruned model and computing a fitness score $f(\mathbf{L})$. Depending on the model, we use one of several lightweight metrics, non-reference measures such as CLIP-IQA~\cite{wang2023exploring} or TOPIQ~\cite{chen2024topiq}, or reference-based SSIM~\cite{wang2004image}—to assess perceptual and semantic fidelity without incurring heavy distribution-level computation.

\paragraph{\textbf{Level-Switch Mutation.}}
To efficiently explore the discrete search space, we apply a level-switch mutation that adaptively redistributes sparsity among diffusion stages while preserving the global sparsity budget.
Given a sparsity schedule $\mathbf{L}$, two stages $(i, j)$ are randomly selected, and their sparsity levels are updated as:
\[
L_i \leftarrow L_i + \Delta, \quad L_j \leftarrow L_j - \Delta,
\]
where $\Delta$ is an integer mutation step drawn from a bounded range,
\[
\Delta \sim \mathcal{U}(1, L_{\text{max}}^{\text{mut}}),
\]
and $L_{\text{max}}^{\text{mut}}$ denotes the maximum mutation magnitude. 
This operation increases the pruning strength of one stage while reducing that of another, ensuring that
\[
\sum_{i=1}^{n} L_i \equiv B,\qquad B := nL_t .
\]

The level-switch mutation serves to maintain a fixed overall sparsity level while allowing controlled redistribution of pruning intensity among diffusion stages. 

\paragraph{\textbf{Selection and Evolution.}}
Following fitness evaluation, candidates are ranked and the top-performing individuals are retained as parents. The level-switch mutation is then applied to these selected candidates to produce offspring for the next generation. This process repeats for a fixed number of iterations or until convergence, yielding an optimized sparsity schedule $\hat{\mathbf{L}}$ that achieves the best trade-off between generative quality and compression ratio. 

\subsection{Stage-wise Second-Order Structural Pruning}
\paragraph{\textbf{SNR-Aware Stage Calibration.}}
To enable stage-specific pruning, following \cite{guo2025mosaicdiff} we construct an \emph{SNR-aware calibration set} that aligns each diffusion stage with its characteristic signal-to-noise ratio (SNR) regime. 
This ensures that each stage is calibrated using latent samples reflecting its true operating noise conditions during inference. Given a standard calibration dataset (e.g., ImageNet-1K \cite{deng2009imagenet}), images are first resized and encoded into latent representations. 
For each diffusion stage, a timestep $t$ is randomly drawn from its interval, and Gaussian noise is applied following the original diffusion noise schedule, ensuring that the resulting latent reproduces the $\text{SNR}(t)$ defined by the model. 
Each tuple consisting of the noised latent, its corresponding timestep 
$t$, and the associated label or caption is collected as calibration data for that stage.
This process provides stage-aligned latent distributions and enables accurate estimation of the local curvature (Hessian) for subsequent pruning. 
By capturing stage-specific denoising behavior, the calibration data supports precise, SNR-consistent pruning decisions across the diffusion process. We emphasize that the calibration only involves adding SNR-matched Gaussian noise at each stage to a standard calibration dataset, and does not require any additional data collection. 

\paragraph{\textbf{Second-Order Structural Pruning.}}
\label{sec:stagewise_secondorder}
Using the SNR-aware calibration data, we perform second-order structured pruning which is an extension of Optimal Brain Surgeon (OBS) to remove redundant parameters while preserving stage-wise representational fidelity. 

For each diffusion stage, given calibration inputs $\mathbf{X}$ and corresponding weights $\mathbf{W}$, we seek a pruned version $\hat{\mathbf{W}}$ that minimizes the deviation between the outputs of the original and pruned weights:
\begin{equation}
    \textstyle \min_{\hat{\mathbf{W}}} \;
    \| \mathbf{W}\mathbf{X} - \hat{\mathbf{W}}_{:, \mathcal{M}}\mathbf{X} \|_2,
    \label{eq:stage-prune}
\end{equation}
where $\mathcal{M}$ denotes a structured pruning mask. 
To account for inter-weight dependencies, the local loss curvature is approximated using the Hessian matrix $\mathbf{H} = \mathbf{X}\mathbf{X}^\top$. 
The second-order importance of each structure is then estimated:
\begin{equation}
    \mathcal{I}_{\mathcal{M}} =
    \textstyle \sum_{i=1}^{d_\text{row}}
    \mathbf{W}_{i,\mathcal{M}}
    \left[
    \left( (\mathbf{H}^{-1})_{\mathcal{M},\mathcal{M}} \right)^{-1}
    \right]
    \mathbf{W}_{i,\mathcal{M}}^{\top},
    \label{eq:importance}
\end{equation}
where structures with smaller $\mathcal{I}_{\mathcal{M}}$ values are considered less critical and thus pruned first.

Following the OBS paradigm, pruning proceeds in a \emph{greedy iterative manner}: at each iteration, the least important column (or structure) is removed, and the remaining weights are updated to compensate for the induced error. 
This compensatory adjustment is computed as:
\begin{equation}
    \boldsymbol{\delta} = \textstyle
    - \mathbf{W}_{:, \mathcal{M}}
    \left[
    \left( (\mathbf{H}^{-1})_{\mathcal{M},\mathcal{M}} \right)^{-1}
    \right]
    \mathbf{H}^{-1}_{\mathcal{M},:}.
    \label{eq:delta-update}
\end{equation}
The pruning–update cycle is repeated until the stage reaches its assigned sparsity level. By explicitly modeling the local curvature and performing iterative OBS-based refinement, this method enables a data-driven, stage-aligned compression process. 
It ensures that pruning decisions respect the intrinsic importance of each structure, providing well-calibrated, second-order pruned models that serve as high-quality candidates for the evolutionary sparsity search. Details about layerdropping and wanda can be found in \Cref{sec:supp_layerdrop_wanda}.

\paragraph{\textbf{Weight Routing Mechanism.}}
Although second-order structured pruning provides high-fidelity, stage-specific compression, it is computationally expensive and memory-intensive, making it difficult to integrate directly with evolutionary search. Two main challenges arise: (\emph{i}) second-order pruning proceeds in a greedy, iterative manner in which each pruning step depends on previously updated weights; executing this procedure for every offspring would require recomputing the entire pruning trajectory repeatedly, yielding prohibitive time costs; and (\emph{ii}) each iteration updates model parameters, so storing multiple stage-specific models, as in MosaicDiff~\cite{guo2025mosaicdiff}, would lead to excessive GPU memory usage when the number of stages grows.

To address these limitations, we introduce a lightweight and memory-efficient \emph{weight routing} mechanism that enables second-order pruning to work seamlessly with evolutionary optimization. The key idea is to precompute the full second-order pruning trajectory for each diffusion stage and store the corresponding updated projection weights. Specifically, we perform a one-time, complete second-order pruning procedure for each stage independently and save the updated weights obtained at each step. This precomputation converts the inherently sequential and time-consuming second-order pruning procedure into a fast, on-demand lookup during search, eliminating redundant computation. Since second-order pruning only modifies the projection layers of the Attention and MLP modules, we maintain a single shared model backbone in GPU memory and dynamically route the appropriate stage-specific projection weights from the precomputed repository as needed. This design enables efficient weight swapping across stages while keeping memory overhead low and maintaining a consistent model structure. During evolutionary search, each offspring directly queries the precomputed database via the weight routing mechanism to retrieve the exact set of stage-specific weights dictated by its sparsity schedule. This eliminates the need for on-the-fly second-order updates, greatly accelerates candidate evaluation, and ensures consistent comparison under identical pruning orders. Overall, weight routing forms an effective bridge between high-precision second-order pruning and scalable evolutionary search, delivering practical acceleration. Loading memory is compared against model stitching in \Cref{tab:memory_compare}. 

%% file: sec/4_experiments.tex
\section{Experiments}
\paragraph{\textbf{Models and Baselines.}}
We evaluate our method on two representative latent diffusion models spanning both U\hbox{-}Net-based~\cite{ronneberger2015u} and transformer-based~\cite{vaswani2017attention} designs, enabling a comprehensive assessment across distinct architectures. 
(1) \textbf{DiT}~\cite{peebles2023scalable} is a transformer-based diffusion model with a pure Vision Transformer backbone, offering strong scalability and competitive image generation quality. 
We focus on pruning the DiT-XL/2 variant trained on $256{\times}256$ images, which contains 675M parameters.
(2) \textbf{SDXL-base-1.0}~\cite{podell2023sdxl} is a U\hbox{-}Net-based latent diffusion model for text-to-image synthesis, featuring 2.6B parameters and high visual fidelity across diverse prompts with a native resolution of $1024{\times}1024$. For baselines, we compare our method with a range of state-of-the-art post-training acceleration approaches for diffusion models. On DiT, we compare our method with Diff-Pruning and MosaicDiff using DDIM sampling~\cite{song2020denoising}. On SDXL, we compare our method with the training-free caching approach DeepCache~\cite{ma2024deepcache}, the first-order structural pruning method Diff-Pruning~\cite{fang2023structural}, the timestep-aware second-order OBS-Diff~\cite{zhu2025obs} with the structural setting, and MosaicDiff~\cite{guo2025mosaicdiff} with Layer Dropping, Structural Wanda~\cite{sun2023simple}, and Second-order Structural Pruning (OBS in short)~\cite{frantar2022optimal}, all using Euler Ancestral Sampling~\cite{karras2022elucidating}. We also test on FLUX.1-schnell\cite{flux2024, labs2025flux1kontextflowmatching} and observe consistent gains over MosaicDiff; results are reported in \Cref{sec:supp_flux}.

\paragraph{\textbf{Dataset and Evaluation Metrics.}}
Following the setup of \cite{guo2025mosaicdiff}, we calibrate SDXL on the MS-COCO 2017 \cite{lin2014microsoft} dataset and DiT on the ImageNet-1K \cite{deng2009imagenet} dataset at a resolution of $256{\times}256$ with 1024 samples.
For SDXL, we generate 5K samples ($1024{\times}1024$) using prompts from the MS-COCO2017 evaluation set and report Fréchet Inception Distance (FID)\cite{heusel2017gans} to assess overall image quality and distributional fidelity, CLIP\text{-}Score~\cite{hessel2021clipscore} (using ViT\text{-}B/16) to evaluate text–image alignment, and SSIM~\cite{wang2004image} to measure structural similarity with the original model outputs.
For DiT, we generate 50K images and compute the FID, Inception Score (IS), Precision, and Recall using the ADM TensorFlow evaluation suite\cite{dhariwal2021diffusion}.

\paragraph{\textbf{Implementation Details.}} In each generation, we maintain 20 individuals composed of 16 offsprings and 4 survivors. Offsprings are generated from one randomly chosen survivor, and all survivors are carried over to the next generation. The search proceeds for 100 generations. For DiT results in \Cref{tab:ditxl2_comparison}, we use 20 stages for Layer dropping and Wanda, 10 stages for OBS and compute fitness with 64 fixed random latents and class labels during searching. Fitness for pruning methods are measured using TOPIQ~\cite{chen2024topiq}. For Layerdrop, the sparsity level is set to 28, aligned with the 28 transformer blocks in DiT-XL/2. For Wanda and OBS, we set the sparsity level to 16. The maximum mutation magnitude is set to 5, each individual undergoes a single mutation per generation. Classifier-free guidance (CFG) for all is set to 2.1. For SDXL results in \Cref{tab:sdxlbase}, during evolutionary search, we use 10 stages and evaluate fitness using 4 fixed prompts from the MS-COCO training set paired with 4 fixed latents. In the Layerdrop setting, the total sparsity level is set to 70, corresponding to SDXL’s 70 transformer blocks in total. For Wanda and OBS, the sparsity level is set to 10, matching the number of attention heads (10 or 20) in each transformer block. We adopt CLIP-IQA~\cite{wang2023exploring} as the fitness function for Layerdrop, Wanda and OBS 30\%. On OBS 10\% and 20\%, we use SSIM computed against the full model’s sampling results. We set the maximum mutation magnitude to approximately 30\% of the maximum sparsity level, corresponding to 20 for LayerDrop and 3 for Wanda and OBS. For a fair comparison, CFG is set to 7.5 across all methods. All experiments are conducted on a single NVIDIA L40S 48GB GPU.

\subsection{Main Results}

\input{tables/table_main_dit}

\input{tables/table_main_sdxl}

\paragraph{\textbf{Results on DiT.}} As shown in \Cref{tab:ditxl2_comparison}, \sysname delivers consistently superior generation quality across all pruning strategies. Under Wanda, \sysname preserves semantic fidelity and texture details significantly better than MosaicDiff, achieving large gains in IS and notable reductions in FID at matched sparsity levels. \sysname performs significantly better in high-sparsity level setting, for example, with sparsity 43.75\% \sysname maintains FID 4.25.

\paragraph{\textbf{Results on SDXL.}} As shown in \Cref{tab:sdxlbase}, our \sysname consistently outperforms baseline pruning methods across all strategies and sparsity levels on SDXL-Base-1.0. Compared with MosaicDiff, \sysname achieves notably lower FID and higher SSIM, indicating superior visual fidelity and structural preservation. Under LayerDrop and OBS pruning, for example, \sysname reduces FID by over 30\% and improves SSIM by up to 0.15. Even at 30\% sparsity, it maintains competitive quality, demonstrating strong robustness to aggressive compression. Overall, \sysname achieves comparable 
performance than dense models while offering substantial parameter efficiency.

\subsection{Analysis}
\begin{figure}[t]
  \centering
  \subfloat[DiT]{%
    \includegraphics[width=0.49\linewidth]{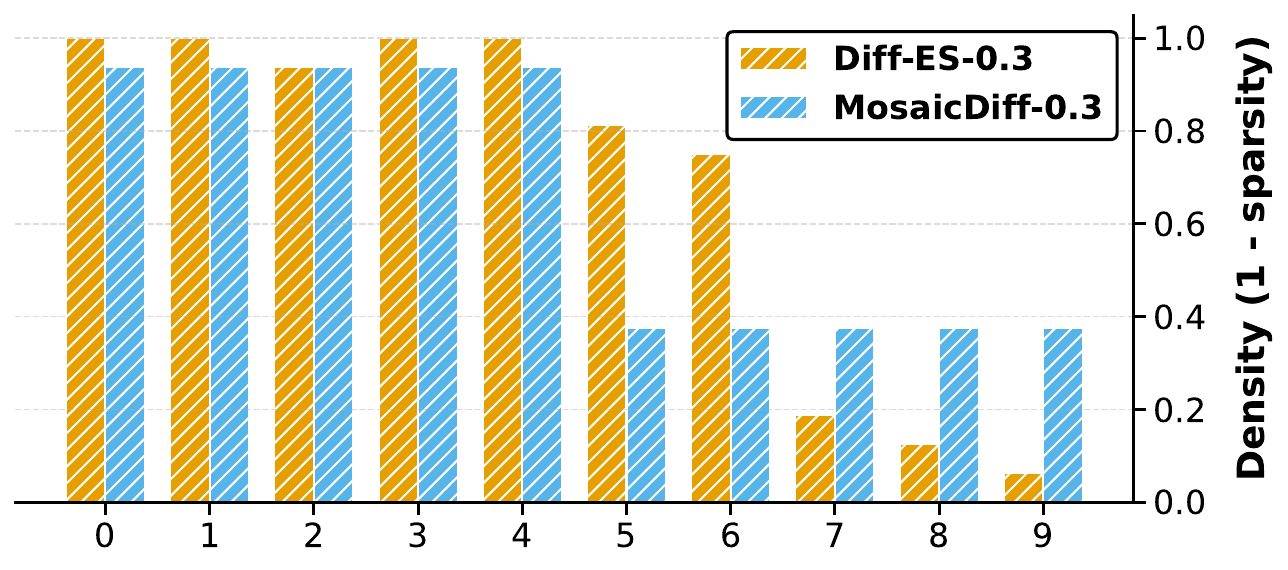}
  }\hfill
  \subfloat[SDXL]{%
    \includegraphics[width=0.49\linewidth]{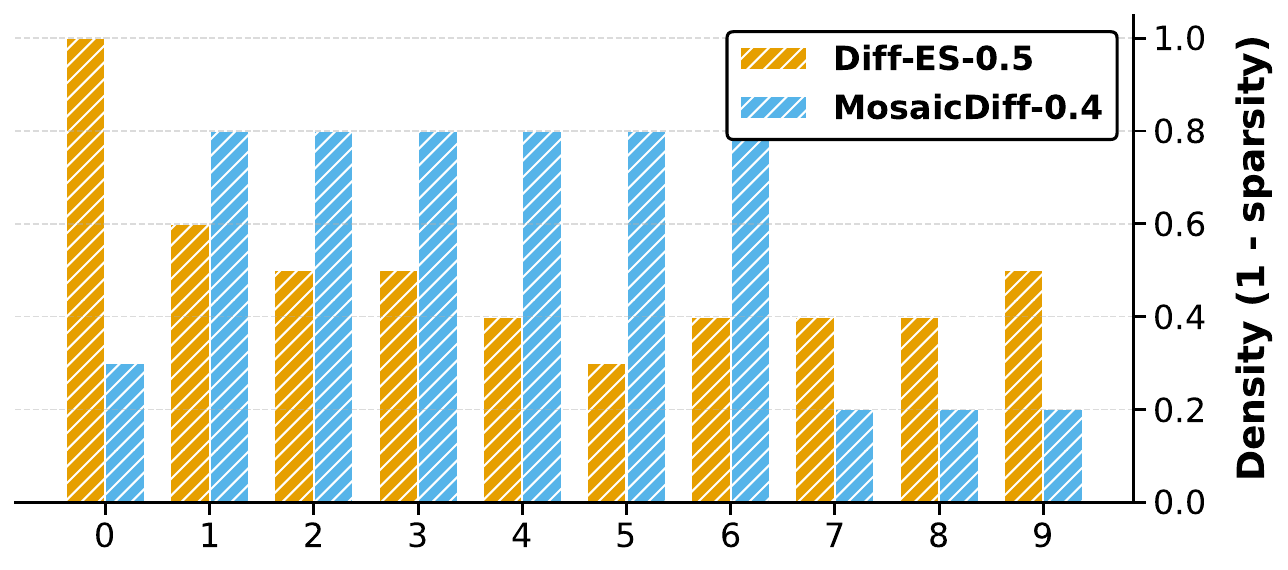}
  }
  \vspace{-0.6em}
  \caption{
  \textbf{Stage-wise sparsity schedules of \sysname and MosaicDiff across models.}
  \textbf{DiT:} \sysname vs MosaicDiff (FID: 3.63 vs 4.01);
  \textbf{SDXL:} \sysname vs MosaicDiff (FID: 33.12 vs 98.56).
  On DiT, the sparsity schedules are similar between MosaicDiff and \sysname.
  On SDXL, the sparsity schedules diverge, indicating that MosaicDiff's empirical schedule overfits to DiT.
  Stages 0--9 correspond to timesteps 0--999; sampling starts at timestep 999, and earlier stages correspond to earlier starting timesteps.
  }
  \label{fig:sparsity_distribution_analysis}
  \vspace{-0.8em}
\end{figure}

\paragraph{\textbf{Sparsity Schedule Analysis.}}
\Cref{fig:sparsity_distribution_analysis} compares the searched stage-wise sparsity schedules of \sysname and MosaicDiff for both SDXL and DiT. MosaicDiff follows an empirically motivated rule: 
the assumed importance ordering of sampling steps is \emph{beginning}~$<$~\emph{end}~$<$~\emph{middle}.
On DiT, our searched stage-wise sparsity schedule exhibits a trend broadly consistent with MosaicDiff's intuition: the middle and late steps are assigned higher density, while pruning is concentrated in earlier stages. This aligns with the fact that DiT is relatively under-trained; degrading the middle or end of the sampling trajectory severely harms global structure and details.
In contrast, SDXL shows a different pattern. \sysname assigns higher density to early and late stages, with moderate density in the middle, which is nearly opposite to MosaicDiff's empirical heuristic. Correspondingly, MosaicDiff's fixed computation allocation leads to noticeably degraded performance on SDXL: even at a lower sparsity level (0.4), its FID is worse than \sysname at sparsity 0.5. These observations suggest that MosaicDiff’s heuristic schedule overfits to DiT-style behavior, while \sysname adapts sparsity to different models and generalizes better.

\paragraph{\textbf{Generalization Beyond Search Set.}}
An important practical question is whether the searched schedule overfits to the limited prompt set used for optimization. Our empirical results indicate that overfitting does not occur.
On SDXL, reducing the search budget from 4 prompts to a single prompt-latent pair (Samp=1 in \Cref{tab:search_cost_scaling_dit_sdxl_compact}) still yields FID 25.75 and CLIP 0.3223, essentially unchanged from the default setting.
\Cref{sec:supp_visual} provides results on AI-generated prompts outside the MS-COCO dataset, which further corroborates this finding, showing that image quality remains stable under the learned schedule on out-of-distribution text inputs.

\input{tables/resource_table}
\paragraph{\textbf{Overall Consumption.}}
\Cref{tab:efficiency_metrics} shows that \sysname uses significantly less average memory than MosaicDiff by routing stage-specific weights within a single model instead of maintaining multiple stitched models in GPU memory. This advantage is more pronounced on larger backbones such as SDXL. On DiT, calibration is relatively cheap, so evolutionary search dominates the total pruning time. On SDXL, calibration is much more expensive, and evolutionary search can compensate for reduced calibration samples while preserving quality. A detailed latency-scaling analysis across batch sizes is in \Cref{sec:supp_latency}.

\subsection{Ablation Study}
\paragraph{\textbf{Fitness.}} \Cref{tab:ablation_table} examines the impact of different lightweight fitness metrics used during evolutionary search.
CLIP-IQA yields the best FID, while SSIM-guided search preserves the most structural details (highest SSIM) with competitive FID.
The metric choice therefore controls the quality trade-off, and \sysname adapts reliably to different optimization objectives.
A full sparsity sweep from 10\% to 50\% for different fitness metrics is provided in \Cref{sec:supp_sdxl}, confirming the same trend beyond the 30\% setting in the main paper. We also compare greedy search with our evolutionary search in \Cref{sec:supp_greedy}.
\paragraph{\textbf{Hyperparameter Sensitivity.}}
\input{tables/ablation_table}
In \Cref{tab:search_cost_scaling_dit_sdxl_compact}, \sysname is generally stable around the default settings. Varying generations has limited impact on final quality: on DiT, reducing Gen from 100 to 50 only slightly changes FID (12.86$\rightarrow$13.41), while Gen=150 brings no further gain; on SDXL, Gen in \{50,100,150\} yields nearly identical CLIP/FID.
Sensitivity is more pronounced for stage granularity on DiT: coarse partitioning (Stage=3/5) significantly degrades quality (FID 19.43/14.48), whereas finer partitioning (Stage=20) improves it to 8.96.
For SDXL, performance is comparatively robust across Samples and Stages (CLIP $\approx$ 0.322, FID mostly around 25.8), indicating a flatter search landscape.
Overall, our default configurations \((100,64,10)\) for DiT and \((100,4,10)\) for SDXL achieve a strong balance between quality and efficiency. And comparing with MosaicDiff (FID 22.29 on DiT at 50\% sparsity and 59.09 on SDXL at 30\% sparsity), \sysname achieves better performance without requiring careful hyperparameter tuning.  

\input{tables/ablation_search_table}

\paragraph{\textbf{Search-cost Scaling.}}
Search cost scales in a predictable manner with the three hyperparameters \((\text{Gen}, \text{Samp}, \text{Stage})\). Increasing \textbf{Gen} mainly increases wall-clock time while leaving peak memory nearly unchanged (DiT: 10.39h at Gen=50 vs 24.99h at Gen=150, both 37.16GB). Increasing \textbf{Samp} primarily raises peak memory on both models (DiT: 22.60GB at Samp=32 vs 41.90GB at Samp=128; SDXL: 14.10GB at Samp=1 vs 43.12GB at Samp=8), with comparatively smaller effect on quality. Among the three knobs, \textbf{Stage} is the strongest runtime driver: DiT increases from 0.99h (Stage=3) to 34.85h (Stage=20), and SDXL from 0.13h to 4.09h. These trends suggest a practical tuning rule: use fewer stages/samples for fast search, and increase stage granularity when maximum generation quality is desired. Notably, even with only 0.13h of search on SDXL (Stage=3), \sysname 
achieves FID 25.77 and CLIP 0.3220, substantially outperforming MosaicDiff at the same 30\% OBS sparsity in \Cref{tab:sdxlbase} (FID 59.09, CLIP 0.3020).

\input{tables/memory_compare} 
\paragraph{\textbf{Combine With Other Acceleration Methods.}}
Because \sysname\ focuses on reducing per-step model computation, it is naturally complementary to step-reduction and feature-reuse accelerators. We therefore combine it with DPM-Solver and DeepCache, and observe additional efficiency gains with only minor quality changes. Detailed quantitative comparisons are provided in \Cref{sec:supp_sdxl}. These results suggest that \sysname\ can be integrated into existing acceleration pipelines with minimal additional engineering overhead.

%% file: tables/table_main_dit.tex
\begin{table*}[t]
\centering
\caption{Result comparison of baselines and \textbf{\sysname{} (Ours)} on  DiT-XL/2. All results are reported using 20 sampling steps for a fair comparison. \sysname{} achieves the best performance compared with all baselines including MosaicDiff on different pruning strategies, demonstrating the effectiveness and generalizability of our method.}
\vspace{-3mm}
\label{tab:ditxl2_comparison}
\setlength{\tabcolsep}{4pt}
\renewcommand{\arraystretch}{1.15}
\resizebox{\linewidth}{!}{
\begin{tabular}{c|c|c|ccc|cccc}
\toprule
\textbf{Pruning Strategy} & \textbf{Method} & \textbf{Sparsity (\%)} & \textbf{MACs} $\downarrow$ & \textbf{Latency (ms)} $\downarrow$ & \textbf{Speedup} $\uparrow$ & \textbf{IS} $\uparrow$ & \textbf{FID} $\downarrow$ & \textbf{Prec.} $\uparrow$ & \textbf{Rec.} $\uparrow$ \\
\midrule

\multirow{1}{*}{Dense} 
& Original & 0\% & 2.37 & 83.60 & 1.00$\times$ & 329.41 & 4.49 & 0.87 & 0.48 \\
\cmidrule(lr){1-10}

\multirow{1}{*}{Diff-Pruning} 
& Original & 20\% & 1.90 & 81.72 & 1.03$\times$ & 156.36 & 15.52 & 0.73 & 0.27 \\
\cmidrule(lr){1-10}

\multirow{2}{*}{\centering LayerDrop}
& MosaicDiff & 17.86\% & 2.02 & 73.34 & 1.14$\times$ & 1.50 & 366.96 & 0.00 & 0.00 \\
& \ourscell{\sysname{} (Ours)} & \ourscell{17.80\%} & \ourscell{2.02} & \ourscell{71.22} & \ourscell{1.17$\times$} & \ourscell{\textbf{36.30}} & \ourscell{\textbf{84.13}} & \ourscell{\textbf{0.22}} & \ourscell{\textbf{0.40}} \\
\cmidrule(lr){1-10}

\multirow{4}{*}{\centering Wanda} 
& MosaicDiff & 12.50\% & 2.13 & 87.13 & 0.96$\times$ & 66.57 & 43.89 & 0.39 & 0.46 \\
& \ourscell{\sysname{} (Ours)} & \ourscell{12.50\%} & \ourscell{2.13} & \ourscell{78.87} & \ourscell{1.06$\times$} & \ourscell{\textbf{253.79}} & \ourscell{\textbf{5.28}} & \ourscell{\textbf{0.79}} & \ourscell{\textbf{0.50}} \\
\cmidrule(lr){2-10}
& MosaicDiff & 18.75\% & 2.02 & 81.73 & 1.03$\times$ & 7.30 & 134.60 & 0.10 & 0.33 \\
& \ourscell{\sysname{} (Ours)} & \ourscell{18.75\%} & \ourscell{2.02} & \ourscell{76.97} & \ourscell{1.09$\times$} & \ourscell{\textbf{17.55}} & \ourscell{\textbf{121.33}} & \ourscell{\textbf{0.25}} & \ourscell{\textbf{0.39}} \\
\cmidrule(lr){1-10}

\multirow{6}{*}{\centering OBS} 
& MosaicDiff & 31.46\% & 1.66 & 68.91 & 1.22$\times$ & 251.23 & 4.01 & 0.79 & \textbf{0.53} \\
& \ourscell{\sysname{} (Ours)} & \ourscell{31.25\%} & \ourscell{1.66} & \ourscell{69.42} & \ourscell{1.20$\times$} & \ourscell{\textbf{268.66}} & \ourscell{\textbf{3.63}} & \ourscell{\textbf{0.81}} & \ourscell{0.52} \\
\cmidrule(lr){2-10}
& MosaicDiff & 42.50\% & 1.42 & 60.96 & 1.38$\times$ & 206.84 & 6.13 & 0.73 & 0.56 \\
& \ourscell{\sysname{} (Ours)} & \ourscell{43.75\%} & \ourscell{1.42} & \ourscell{61.20} & \ourscell{1.37$\times$} & \ourscell{\textbf{232.73}} & \ourscell{\textbf{4.25}} & \ourscell{\textbf{0.77}} & \ourscell{\textbf{0.57}} \\
\cmidrule(lr){2-10}
& MosaicDiff & 50.00\% & 1.19 & 55.25 & 1.52$\times$ & 110.13 & 22.29 & 0.55 & 0.58 \\
& \ourscell{\sysname{} (Ours)} & \ourscell{50.00\%} & \ourscell{1.19} & \ourscell{56.92} & \ourscell{1.47$\times$} & \ourscell{\textbf{156.23}} & \ourscell{\textbf{12.86}} & \ourscell{\textbf{0.66}} & \ourscell{\textbf{0.59}} \\
\bottomrule
\end{tabular}
}
\end{table*}

%% file: tables/table_main_sdxl.tex
\begin{table}[t]
  \centering
  \caption{Result comparison of baselines and \textbf{\sysname{} (Ours)} on SDXL-Base-1.0 with 20 sampling steps. Our method consistently achieves better perceptual quality across pruning strategies and sparsity levels, with particularly strong gains at higher sparsity.}
  \vspace{-3mm}
  \label{tab:sdxlbase}
  \setlength{\tabcolsep}{6pt}
  \renewcommand{\arraystretch}{1.12}
\resizebox{0.75\linewidth}{!}{
  \begin{tabular}{c|c|c|ccc}
    \toprule
    \textbf{Pruning Strategy} & \textbf{Method} & \textbf{Sparsity (\%)} &
    \textbf{FID} $\downarrow$ & \textbf{CLIP-Score} $\uparrow$ & \textbf{SSIM} $\uparrow$ \\
    \midrule

    Dense & Original & 0\% & 25.46 & 0.3219 & 1.000 \\
    \cmidrule(lr){1-6}

    Deep-Cache & Original & — & 26.32 & 0.3185 & 0.6709 \\
    \cmidrule(lr){1-6}

    Diff-Pruning & Original & 20\% & 96.5 & 0.2702 & 0.5887 \\
    \cmidrule(lr){1-6}

    OBS-Diff & Original & 30\% & 28.49 & 0.3215 & 0.6760 \\
    \cmidrule(lr){1-6}

    \multirow{6}{*}{\centering LayerDrop}
      & MosaicDiff      & 10\% & 25.16 & 0.3198 & 0.6341 \\
      & \ourscell{\sysname{} (Ours)} & \ourscell{10\%} & \ourscell{\textbf{24.12}} & \ourscell{\textbf{0.3214}} & \ourscell{\textbf{0.767}} \\
      \cmidrule{2-6}
      & MosaicDiff      & 20\% & 29.99 & 0.3157 & 0.5687 \\
      & \ourscell{\sysname{} (Ours)} & \ourscell{20\%} & \ourscell{\textbf{24.56}} & \ourscell{\textbf{0.3193}} & \ourscell{\textbf{0.7016}} \\
      \cmidrule{2-6}
      & MosaicDiff      & 30\% & 39.27 & 0.3096 & 0.5302 \\
      & \ourscell{\sysname{} (Ours)} & \ourscell{30\%} & \ourscell{\textbf{29.71}} & \ourscell{\textbf{0.3144}} & \ourscell{\textbf{0.6737}} \\
    \cmidrule(lr){1-6}

    \multirow{6}{*}{\centering Wanda}
      & MosaicDiff      & 10\% & 25.83 & \textbf{0.3226} & 0.6292 \\
      & \ourscell{\sysname{} (Ours)} & \ourscell{10\%} & \ourscell{\textbf{25.62}} & \ourscell{0.3222} & \ourscell{\textbf{0.6457}} \\
      \cmidrule{2-6}
      & MosaicDiff      & 20\% & 31.22 & 0.3183 & 0.5991 \\
      & \ourscell{\sysname{} (Ours)} & \ourscell{20\%} & \ourscell{\textbf{26.99}} & \ourscell{\textbf{0.3203}} & \ourscell{\textbf{0.6317}} \\
      \cmidrule{2-6}
      & MosaicDiff      & 30\% & 37.24 & 0.3111 & 0.5655 \\
      & \ourscell{\sysname{} (Ours)} & \ourscell{30\%} & \ourscell{\textbf{36.88}} & \ourscell{\textbf{0.3139}} & \ourscell{\textbf{0.6179}} \\
    \cmidrule(lr){1-6}

    \multirow{6}{*}{\centering OBS}
      & MosaicDiff      & 10\% & 25.32 & 0.3212 & 0.6352 \\
      & \ourscell{\sysname{} (Ours)} & \ourscell{10\%} & \ourscell{\textbf{24.80}} & \ourscell{\textbf{0.3221}} & \ourscell{\textbf{0.8752}} \\
      \cmidrule{2-6}
      & MosaicDiff      & 20\% & 39.6  & 0.3095 & 0.5462 \\
      & \ourscell{\sysname{} (Ours)} & \ourscell{20\%} & \ourscell{\textbf{24.48}} & \ourscell{\textbf{0.3222}} & \ourscell{\textbf{0.8341}} \\
      \cmidrule{2-6}
      & MosaicDiff      & 30\% & 59.09 & 0.3020  & 0.5367 \\
      & \ourscell{\sysname{} (Ours)} & \ourscell{30\%} & \ourscell{\textbf{25.87}} & \ourscell{\textbf{0.3214}} & \ourscell{\textbf{0.7118}} \\
    \bottomrule
  \end{tabular}
  }
\end{table}

%% file: tables/resource_table.tex
\begin{table}[t]
\centering
\caption{
Efficiency and quality comparison on DiT ($\sim$40\% sparsity) and SDXL (30\% sparsity).
Diff-ES achieves strong acceleration while preserving high generative fidelity. 
For latency, we use a batch size of 8. Runtime denotes the one-time computational cost incurred by calibration and evolutionary search.
For SDXL, we use SSIM as the fitness metric and perform a single calibration shared across stages.
The number of generations is set to 30.
}
\vspace{-1em}
\label{tab:efficiency_metrics}
\setlength{\tabcolsep}{2pt}
\renewcommand{\arraystretch}{1.15}
\resizebox{0.722\columnwidth}{!}{
\begin{tabular}{l|l|c|c|c|c}
\toprule
\textbf{Model} & \textbf{Method} & \textbf{Latency (ms)} $\downarrow$ & \textbf{Runtime (h)} $\downarrow$ & \textbf{Speedup} $\uparrow$ & \textbf{FID} $\downarrow$ \\
\midrule
 & Dense & 83.60 & 0 & 1 & 4.49 \\
DiT & MDiff & 60.96 & 0.64 & 1.38 & 6.13 \\
& Diff-ES & 61.20 & 18.87 & 1.37 & 4.25 \\
\midrule
 & Dense & 1627.69 & 0 & 1 & 25.46 \\
SDXL & MDiff & 1379.40 & 6.62 & 1.18 & 59.09 \\
& Diff-ES & 1479.72 & 3.61 & 1.10 & 25.88 \\
\bottomrule
\end{tabular}
}
\end{table}

%% file: tables/ablation_table.tex
\begin{wraptable}{r}{0.5\linewidth}
  \setlength{\intextsep}{0.35em}
  \setlength{\columnsep}{0.8em}
  \vspace{-3.4em}
  \centering
  \caption{Evaluation of \textbf{Diff-ES + OBS} under different fitness metrics on SDXL at 30\% sparsity.}
  \label{tab:ablation_table}
  \setlength{\tabcolsep}{3.5pt}
  \renewcommand{\arraystretch}{1.1}
  \small
  \resizebox{0.7\linewidth}{!}{
  \begin{tabular}{lccc}
    \toprule
    \textbf{Fitness} & \textbf{FID} $\downarrow$ & \textbf{CLIP} $\uparrow$ & \textbf{SSIM} $\uparrow$ \\
    \midrule
    CLIP-IQA & 25.87 & 0.3214 & 0.7118 \\
    TOPIQ    & 26.04 & 0.3206 & 0.6777 \\
    SSIM     & 26.03 & 0.3223 & 0.8220 \\
    \bottomrule
  \end{tabular}}
  \vspace{-2em}
\end{wraptable}

%% file: tables/ablation_search_table.tex
\begin{table*}[tb]
  \centering
  \caption{
  Hyperparameter sensitivity and search-cost scaling of Diff-ES with OBS.
  One factor in $(\text{Gen}, \text{Samp}, \text{Stage})$ is varied at a time.
  For DiT (50\% sparsity), the base hyperparameters $(100, 64, 10)$ are identical to those used in \Cref{tab:ditxl2_comparison}; TOPIQ is adopted as the fitness metric.
  For SDXL (30\% sparsity), the base hyperparameters $(100, 4, 10)$ match those in \Cref{tab:sdxlbase}; SSIM is used as the fitness metric.
  }
  \label{tab:search_cost_scaling_dit_sdxl_compact}
  \vspace{-0.6em}

  \begin{minipage}[t]{0.49\linewidth}
    \centering
    \small
    DiT + OBS (50\%)
    \resizebox{\linewidth}{!}{%
    \begin{tabular}{l c c c c}
      \toprule
      \textbf{Setting} & \textbf{IS} $\uparrow$ & \textbf{FID} $\downarrow$ & \textbf{Time (h)} $\downarrow$ & \textbf{Mem (GB)} $\downarrow$ \\
      \midrule
      base    & {156.23} & {12.86} & {17.26} & {37.16} \\
      \midrule
      Gen = 50      & 152.39 & 13.41 & 10.39 & 37.16 \\
      Gen = 150     & 156.23 & 12.86 & 24.99 & 37.16 \\
      \addlinespace
      Samp = 32     & 154.84 & 13.04 & 17.62 & \textbf{22.60} \\
      Samp = 128    & 129.46 & 18.29 & 16.28 & 41.90 \\
      \phantom{Samp = 8} & \phantom{\textbf{0.3221}} & \phantom{25.91} & \phantom{2.46} & \phantom{43.12} \\
      \addlinespace
      Stage = 3     & 123.23 & 19.43 & \textbf{0.99} & 34.32 \\
      Stage = 5     & 145.27 & 14.48 & 4.80 & 35.12 \\
      Stage = 20    & \textbf{183.59} & \textbf{8.96} & 34.85 & 41.25 \\
      \bottomrule
    \end{tabular}}
  \end{minipage}
  \hfill
  \begin{minipage}[t]{0.49\linewidth}
    \centering
    \small
    SDXL + OBS (30\%)
    \resizebox{\linewidth}{!}{%
    \begin{tabular}{l c c c c}
      \toprule
      \textbf{Setting} & \textbf{CLIP} $\uparrow$ & \textbf{FID} $\downarrow$ & \textbf{Time (h)} $\downarrow$ & \textbf{Mem (GB)} $\downarrow$ \\
      \midrule
      {base}   & {0.3222} & {25.81} & {1.71} & {22.84} \\
      \midrule
      Gen = 50      & 0.3222 & 25.81 & 1.57 & 22.84 \\
      Gen = 150     & 0.3222 & 25.81 & 1.70 & 22.84 \\
      \addlinespace
      Samp = 1      & 0.3223 & \textbf{25.75} & 0.53 & \textbf{14.10} \\
      Samp = 2      & \textbf{0.3224} & 25.87 & 0.82 & 21.31 \\
      Samp = 8      & 0.3221 & 25.91 & 2.46 & 43.12 \\
      \addlinespace
      Stage = 3     & 0.3220 & 25.77 & \textbf{0.13} & 35.43 \\
      Stage = 5     & 0.3223 & 26.88 & 0.34 & 35.43 \\
      Stage = 20    & 0.3222 & 25.85 & 4.09 & 35.43 \\
      \bottomrule
    \end{tabular}}
  \end{minipage}

  \vspace{-1em}
\end{table*}

%% file: tables/memory_compare.tex
\begin{wraptable}{r}{0.5\linewidth}
  \setlength{\intextsep}{0.35em}
  \setlength{\columnsep}{0.8em}
  \vspace{-3.3em}
  \centering
  \caption{Loading memory comparison of \sysname with model stitching and proposed weight routing. We report results for DiT-XL/2 at $\sim$40\% sparsity, the number of stages is 10. Weight routing saves $\sim$42.7\% GPU memory compared to model stitching.}
  \label{tab:memory_compare}
  \setlength{\tabcolsep}{6pt}
  \renewcommand{\arraystretch}{1.12}
  \small
  \resizebox{0.7\linewidth}{!}{
  \begin{tabular}{l|c}
    \toprule
    \textbf{Setting} & \textbf{Memory (GB)} $\downarrow$ \\
    \midrule
    Model Stitching & 9.79 \\
    Weight Routing  & \textbf{5.61} \\
    \bottomrule
  \end{tabular}}
  \vspace{-3em}
\end{wraptable}

%% file: sec/5_conclusion.tex
\section{Conclusion}
We introduced \sysname{}, a stage-wise structural diffusion pruning framework that leverages evolutionary search to discover an optimal stage-wise sparsity schedule along the denoising trajectory. By coupling global optimization with memory-efficient weight routing, \sysname achieves superior trade-offs between efficiency and fidelity without retraining or model duplication. Extensive experiments on both CNN- and Transformer-based diffusion models demonstrate consistent improvements over prior methods, establishing \sysname as a robust and general solution for efficient diffusion model acceleration.

%% file: sec/6_suppl.tex
\begin{figure*}[t]
    \centering
    \includegraphics[width=1\linewidth]{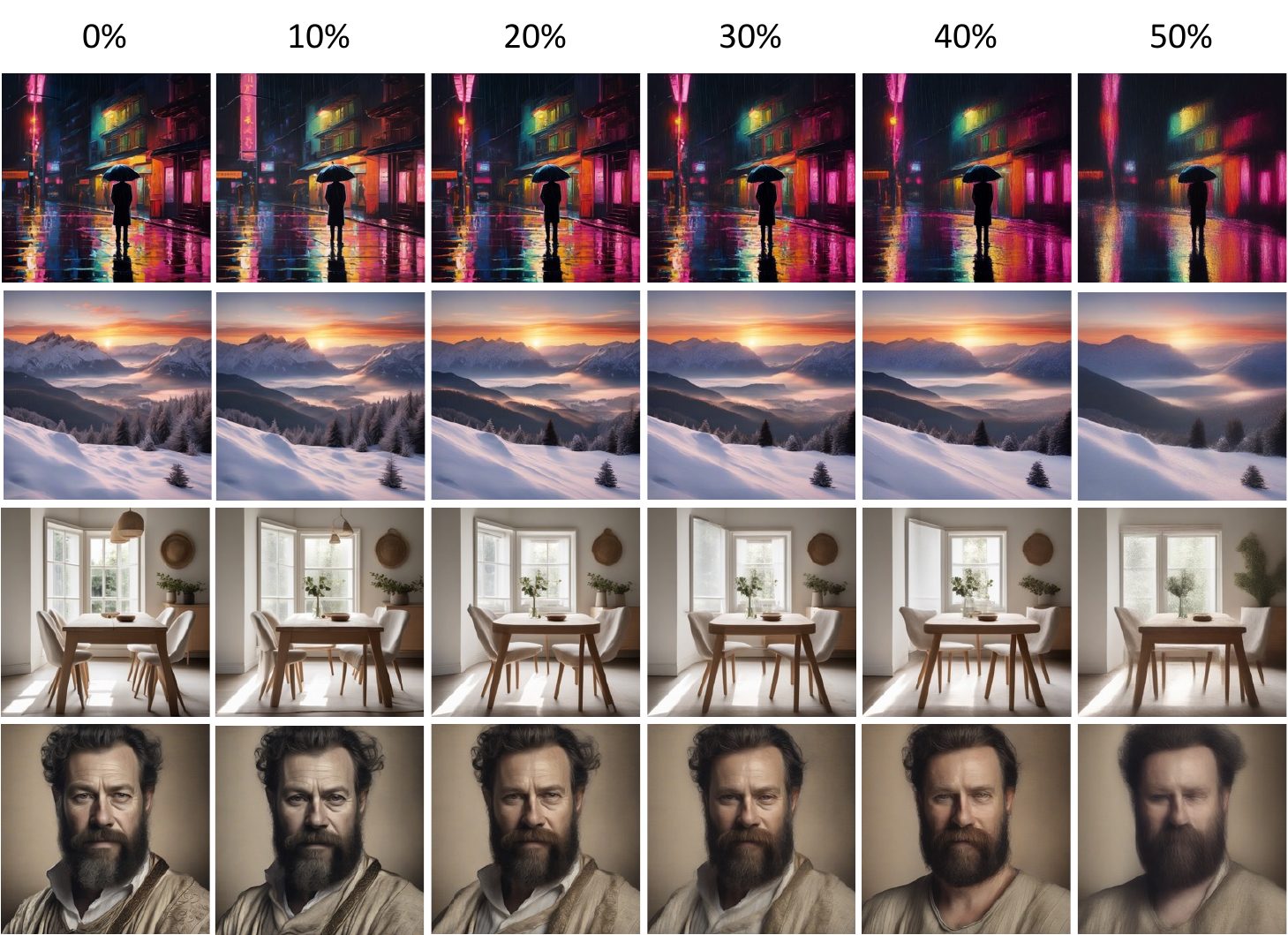}
    \caption{Out-of-distribution visual examples on AI-generated prompts (disjoint from the MS-COCO search set) across sparsity levels under the same fitness metric (CLIP-IQA). \sysname preserves semantic consistency and fine details up to moderate sparsity, with more visible degradation mainly at the highest sparsity level.}
    \label{fig:placeholder}
\end{figure*}

\section{More Experimental Results on SDXL}
\label{sec:supp_sdxl}

\input{tables/sup_compatibility}

\input{tables/sup_table_fitness}
\paragraph{\textbf{Compatibility with Other Acceleration Methods.}}
\Cref{tab:sup_compatibility} shows that \sysname\ is compatible with existing acceleration strategies. At 20 steps on SDXL, \sysname\ reduces MACs from 67.64T (Dense) to 52.45T with only a small FID change (25.46$\rightarrow$25.87). Combining \sysname\ with DeepCache further reduces MACs to 36.05T (FID 27.10), and combining \sysname\ with DPM-Solver at 10 steps reduces MACs to 26.23T (FID 25.67). These results indicate that \sysname\ is orthogonal to step-reduction and feature-reuse methods and has the ability to provide additive efficiency gains.

\paragraph{\textbf{\sysname\ Results on SDXL-Base-1.0.}}
\Cref{tab:sup_fitness} compares different fitness metrics used during our OBS-based pruning process. The results show that the choice of fitness function influences the balance between generative quality and structural sparsity. CLIP-IQA and TOPIQ, which emphasize perceptual alignment, help maintain competitive FID and CLIP-Score even at high sparsity levels (e.g., 40\%), whereas SSIM-driven pruning yields stronger pixel-wise structural preservation, as evidenced by substantially higher SSIM values. Overall, CLIP-IQA exhibits the most consistent robustness across sparsity levels. In contrast, SSIM---achieving the lowest FID and highest SSIM at 10\% and 20\%---is preferable in the low-sparsity regime, while TOPIQ becomes slightly more favorable at the extreme sparsity level of 50\%. These findings validate that our metric-aware design enables flexible quality objectives depending on deployment needs and supports the selection of an appropriate fitness function tailored to downstream requirements.

\section{Additional Visual Examples}
\label{sec:supp_visual}
\Cref{fig:placeholder} presents visualization results of \sysname\ across multiple sparsity levels under the same fitness metric (CLIP-IQA). The prompts in this figure are AI-generated and disjoint from the MS-COCO prompts used during evolutionary search, providing an out-of-distribution test. We use diverse prompts covering \textit{urban night scenes}, \textit{natural landscapes}, \textit{indoor environments}, and \textit{portrait generation} to evaluate robustness across semantic categories and texture structures. As shown, pruning up to 30\% sparsity introduces almost no perceptual degradation, with global appearance, sharpness, and fine textures effectively preserved. Even at 40\% sparsity, structural fidelity remains largely intact; although slight softness may appear in high-frequency regions, overall visual quality continues to satisfy demanding aesthetic requirements. Noticeable degradation only becomes evident at 50\% sparsity, particularly in human facial details and object boundaries. These results demonstrate that \sysname\ can safely remove a substantial portion of model weights while maintaining realistic and coherent image generation---especially valuable for resource-constrained or high-throughput deployment scenarios.

\section{Comparison on FLUX.1-schnell}
\label{sec:supp_flux}
\input{tables/sup_table_flux}

To further evaluate the effectiveness of our method, we conduct additional experiments on FLUX.1-schnell, a 12B-parameter rectified-flow--based diffusion model that can generate images in only four sampling steps. This experiment confirms that \sysname\ scales to large models, remains effective in the few-step regime, and naturally extends to rectified-flow--based diffusion models. For the evolutionary search, we use 16 offspring and 4 survivors, with 4 stages and 4 fixed latent codes for fitness evaluation. We run 50 generations and adopt SSIM as the fitness metric. Since the original MosaicDiff work does not report results on FLUX, we reimplement it by following the principle in their paper, where the importance of sampling steps is ordered as \emph{beginning}~$<$~\emph{end}~$<$~\emph{middle}. Under this design, the per-stage sparsity levels are $(0.12, 0.02, 0.25)$ for an overall sparsity of 20\%, with stage proportions fixed at $(0.25, 0.5, 0.25)$. 
The results in \Cref{tab:sup_flux} show that \sysname\ consistently outperforms MosaicDiff on FLUX.1-schnell, further supporting the effectiveness of our approach.

\section{Details About Layerdropping And Wanda}
\label{sec:supp_layerdrop_wanda}
\paragraph{\textbf{Layer Dropping.}}
For each transformer block $f_\ell$, we measure how much the block changes its input by computing the cosine similarity between its input $\mathbf{x}_\ell$ and output $\mathbf{y}_\ell = f_\ell(\mathbf{x}_\ell)$:
\begin{equation}
    s_\ell =\textstyle
    \frac{
        \langle \operatorname{vec}(\mathbf{x}_\ell),\, \operatorname{vec}(\mathbf{y}_\ell) \rangle
    }{
        \|\operatorname{vec}(\mathbf{x}_\ell)\|_2\,
        \|\operatorname{vec}(\mathbf{y}_\ell)\|_2
    }.
    \label{eq:layerdrop_cos}
\end{equation}
We average $s_\ell$ over the calibration set and timesteps to obtain a redundancy score $\bar{s}_\ell$.  
Blocks with the largest $\bar{s}_\ell$ (i.e., behaving closest to identity) are dropped according to the stage-wise sparsity schedule, replacing $f_\ell(\cdot)$ by a skip connection.

\paragraph{\textbf{Structural Wanda.}}
For a linear layer $\mathbf{W}\in\mathbb{R}^{C_{\text{out}}\times C_{\text{in}}}$ and input activations 
$\mathbf{X}\in\mathbb{R}^{(N\!\times\!L)\times C_{\text{in}}}$, the element-wise Wanda score is defined as
\begin{equation}
    S_{ij} = |W_{ij}| \cdot \|\mathbf{X}_{:,j}\|_2.
\end{equation}
Let $\mathcal{A}$ denote the set of output-channel indices corresponding to one attention head.
The structural importance of that head is computed by averaging the element-wise Wanda scores:
\begin{equation}
   \textstyle S_{\text{attn}}
    =
    \frac{1}{|\mathcal{A}|}
    \sum_{i\in\mathcal{A}}
    \sum_{j=1}^{C_{\text{in}}}
    |W_{ij}| \,\|\mathbf{X}_{:,j}\|_2.
\end{equation}
Let $\mathcal{C}$ denote the row index corresponding to one MLP expansion channel in the first FC layer.
Its structural score is obtained similarly by an average:
\begin{equation}
    S_{\text{mlp}}
    =
   \textstyle \frac{1}{C_{\text{in}}}
    \sum_{j=1}^{C_{\text{in}}}
    |W_{\mathcal{C},j}|\,\|\mathbf{X}_{:,j}\|_2.
\end{equation}

Structures with the smallest scores are pruned according to the stage-wise sparsity schedule.

\section{Latency Scaling with Batch Size}
\label{sec:supp_latency}
\input{tables/latency_batchsize}
\Cref{tab:dit_latency_batchsize} shows that the practical benefit of structural pruning is strongly batch-size dependent. At batch size 1, the pruned model is slower than the dense baseline, indicating that the extra weight-routing cost outweighs the parameter-reduction gain. At batch size 2, performance is essentially at parity, marking the transition point where pruning overhead is amortized. From batch size 8 onward, the acceleration becomes clear and stable, with the strongest gain at batch size 8 (1.369$\times$), followed by consistent improvements at 16 and 32. This pattern is expected in real deployment: moderate batch sizes provide enough parallel work to expose the reduction in arithmetic workload, while very large batches gradually shift the bottleneck toward memory traffic and scheduling, which narrows the relative speedup. Overall, these results support using \sysname\ in throughput-oriented settings, where the compute reduction translates into tangible wall-clock latency savings.

\section{Why Evolutionary Search Instead of Greedy Search}
\label{sec:supp_greedy}
\input{tables/greedy_search_compare}
\Cref{tab:greedy_search_compare} compares two schedule optimization strategies under the same setting. We make sure the number of evaluated individuals is the same for both methods. The results show that evolutionary search achieves better performance than greedy search. We suspect greedy search is more prone to getting stuck in a local optimum.

%% file: tables/sup_compatibility.tex
\begin{table}[t]
\centering
\caption{Compatibility of \sysname\ with different acceleration and sampling techniques on SDXL. The table demonstrates that \sysname\ seamlessly integrates with existing methods such as DeepCache and DPM-Solver, maintaining competitive FID while significantly reducing computation. We use \sysname-0.3 and DeepCache-N=2 here. The fitness metric for \sysname is CLIP-IQA.}
\label{tab:sup_compatibility}
\setlength{\tabcolsep}{8pt}
\renewcommand{\arraystretch}{1.15}
\resizebox{0.75\columnwidth}{!}{
\begin{tabular}{l|c|c|c}
\toprule
\textbf{Method} & \textbf{Steps} & \textbf{MACs (T)} & \textbf{FID} $\downarrow$ \\
\midrule
\multicolumn{4}{c}{\textbf{Low-step comparison}} \\
\midrule
Dense                     & 20 & 67.64 & 25.46 \\
\sysname                  & 20 & 52.45 & 25.87 \\
DeepCache + \sysname      & 20 & 36.05 & 27.10 \\
DPM-Solver + \sysname     & 10 & 26.23 & 25.67 \\
\bottomrule
\end{tabular}
}
\end{table}

%% file: tables/sup_table_fitness.tex
\begin{table}[t]
  \centering
  \caption{Comparison of OBS pruning under different fitness metrics on SDXL-Base-1.0.}
  \vspace{-2mm}
  \label{tab:sup_fitness}
  \setlength{\tabcolsep}{6pt}
  \renewcommand{\arraystretch}{1.12}
  \resizebox{1\linewidth}{!}{
  \begin{tabular}{c|c|c|ccc}
    \toprule
    \textbf{Method} & \textbf{Fitness} & \textbf{Sparsity} &
    \textbf{FID} $\downarrow$ & \textbf{CLIP-Score} $\uparrow$ & \textbf{SSIM} $\uparrow$ \\
    \midrule

    Dense & -- & 0\% & 25.46 & 0.3219 & 1.000 \\
    \midrule

    \multirow{5}{*}{\sysname{} (Ours)}
        & \multirow{5}{*}{CLIP-IQA}
        & 10\% & 25.42 & 0.3211 & 0.7725 \\
        & & 20\% & 25.34 & 0.3215 & 0.7146 \\
        & & 30\% & 25.87 & 0.3214 & 0.7118 \\
        & & 40\% & 28.46 & 0.3209 & 0.6980 \\
        & & 50\% & 41.20 & 0.3158 & 0.6994 \\
    \midrule

    \multirow{5}{*}{\sysname{} (Ours)}
        & \multirow{5}{*}{TOPIQ}
        & 10\% & 25.19 & 0.3214 & 0.7500 \\
        & & 20\% & 25.15 & 0.3211 & 0.6990 \\
        & & 30\% & 26.04 & 0.3206 & 0.6777 \\
        & & 40\% & 28.71 & 0.3196 & 0.7043 \\
        & & 50\% & 36.47 & 0.3180 & 0.6769 \\
    \midrule

    \multirow{5}{*}{\sysname{} (Ours)}
        & \multirow{5}{*}{SSIM}
        & 10\% & 24.80 & 0.3221 & 0.8785 \\
        & & 20\% & 24.48 & 0.3222 & 0.8341 \\
        & & 30\% & 26.03 & 0.3223 & 0.8220 \\
        & & 40\% & 28.92 & 0.3208 & 0.7740 \\
        & & 50\% & 42.69 & 0.3154 & 0.7333 \\
    \bottomrule
  \end{tabular}
  }
\end{table}

%% file: tables/sup_table_flux.tex
\begin{table}[t]
  \centering
  \caption{Comparison of MosaicDiff and our \sysname with Layer Dropping on FLUX.1-schnell. The CFG scale is set to 3.5, which is the default value of FLUX.1-schnell.}
  \vspace{-2mm}
  \label{tab:sup_flux}
  \setlength{\tabcolsep}{6pt}
  \renewcommand{\arraystretch}{1.12}
  \resizebox{1\linewidth}{!}{
  \begin{tabular}{c|c|ccc}
    \toprule
    \textbf{Method} & \textbf{Sparsity} &
    \textbf{FID} $\downarrow$ & \textbf{CLIP-Score} $\uparrow$ & \textbf{SSIM} $\uparrow$ \\
    \midrule
    Dense & 0\%  & 31.16 & 0.3187 & 1.0000 \\
    \midrule
    MosaicDiff & 20\% & 30.16 & 0.3084 & 0.4602 \\
    \textbf{\sysname (Ours)} & \textbf{20\%} & \textbf{25.88} & \textbf{0.3150} & \textbf{0.6782} \\
    \bottomrule
  \end{tabular}
  }
\end{table}

%% file: tables/latency_batchsize.tex
\begin{table}[t]
  \centering
  \caption{Overall latency and speedup on DiT with 43.75\% ($\sim$ 40\%) sparsity under different batch sizes. Latency is reported in milliseconds (ms), and speedup is computed as FULL/PRUNED. }
  \label{tab:dit_latency_batchsize}
  \setlength{\tabcolsep}{5pt}
  \renewcommand{\arraystretch}{1.1}
  \resizebox{0.8\columnwidth}{!}{
  \begin{tabular}{c|c|c|c}
    \toprule
    \textbf{Batch Size} & \textbf{FULL (ms)} $\downarrow$ & \textbf{PRUNED (ms)} $\downarrow$ & \textbf{Speedup} $\uparrow$ \\
    \midrule
    1  & 158.7  & 188.5  & 0.842$\times$ \\
    2  & 201.5  & 200.9  & 1.003$\times$ \\
    8  & 670.3  & 489.6  & \textbf{1.369$\times$} \\
    16 & 1320.9 & 1021.6 & 1.293$\times$ \\
    32 & 2877.4 & 2366.3 & 1.216$\times$ \\
    \bottomrule
  \end{tabular}}
\end{table}

%% file: tables/greedy_search_compare.tex
\begin{table}[t]
  \centering
  \caption{Comparison between evolutionary search and greedy search on DiT-XL/2 with OBS pruning at 50\% sparsity with fitness TOPIQ. The number of stages is 10.}
  \label{tab:greedy_search_compare}
  \setlength{\tabcolsep}{6pt}
  \renewcommand{\arraystretch}{1.12}
  \small
  \begin{tabular}{lcccc}
    \toprule
    \textbf{Search Method} & \textbf{IS} $\uparrow$ & \textbf{FID} $\downarrow$ & \textbf{Prec.} $\uparrow$ & \textbf{Rec.} $\uparrow$ \\
    \midrule
    Greedy Search & 83.66 & 33.75 & 0.5028 & 0.5758 \\
    Evolutionary Search (\sysname{}) & \textbf{156.23} & \textbf{12.86} & \textbf{0.66} & \textbf{0.59} \\
    \bottomrule
  \end{tabular}
\end{table}